\documentclass[letterpaper]{article}
\usepackage[preprint]{aaai2027}  
\usepackage[hyphens]{url}
\usepackage{graphicx} 
\usepackage{natbib}
\usepackage{caption}
\usepackage{amsmath}
\usepackage{amssymb}
\usepackage{booktabs}
\usepackage{multirow}
\usepackage{colortbl}
\usepackage[ruled]{algorithm2e}
\usepackage{enumitem}
\usepackage[most]{tcolorbox}

\DontPrintSemicolon
\SetKwInOut{KwIn}{Input}
\SetKwInOut{KwOut}{Output}

\newcommand{\ours}{MGSD}

\title{Learning Visual Spatial Planning from Symbolic State via Modality-Gap-Aware Self-Distillation}
\author{ 
    Haocheng Luo\textsuperscript{\rm 1,3}\equalcontrib, 
    Jiahui Liu\textsuperscript{\rm 1}\equalcontrib, 
    Ruicheng Zhang\textsuperscript{\rm 1,3}, 
    Zhizhou Zhong\textsuperscript{\rm 2}, 
    Jiaqi Huang\textsuperscript{\rm 1}, 
    Zunnan Xu\textsuperscript{\rm 1}, 
    Quan Shi\textsuperscript{\rm 1}, 
    Jun Zhou\textsuperscript{\rm 1}, 
    Shuiyang Mao\textsuperscript{\rm 3}, 
    Wei Liu\textsuperscript{\rm 3},
    Xiu Li\textsuperscript{\rm 1}\corresponding 
}
\affiliations{ 
    \textsuperscript{\rm 1}Tsinghua University\\ 
    \textsuperscript{\rm 2}The Hong Kong University of Science and Technology \\
    \textsuperscript{\rm 3}Video Rebirth
}

\begin{document}
\maketitle

\begin{abstract}
While Vision-Language Models excel at general multimodal understanding, they still struggle with visual spatial planning. We attribute this limitation to a \textit{perception--reasoning modality gap}. Visual planning requires models to infer latent state structures from pixels and then reason over the recovered structure to produce valid actions, whereas symbolic planning directly leverages explicit representation. This discrepancy introduces two sequential bottlenecks: visual state recovery at the perception stage and multi-step planning at the reasoning stage.
To address this, we propose \textbf{\ours{}}, a two-stage modality-gap-aware self-distillation framework. First, a cold-start grounding stage establishes reliable visual state recovery before on-policy training. Second, a symbol-guided on-policy self-distillation stage transfers the privileged teacher's planning behavior to the student through token-level supervision on student-generated prefixes. Crucially, symbolic information is used only during training, while inference relies exclusively on visual inputs. Experiments on visual planning benchmarks show that \ours{} consistently improves performance across different model scales, raising the macro average by 19.3\% and 18.4\%, respectively. The resulting models substantially reduce the gap to the upper bounds obtained with symbolic inputs. Ablation studies and diagnostic analyses further confirm that the gains arise from improvements in both visual state recovery and optimal-path reasoning. These results demonstrate that \textbf{\ours{}} strengthens not only the recovery of actionable states from visual observations but also the ability to plan over the inferred structures. Code is available at \url{https://github.com/Oranger-l/MGSD}.
\end{abstract}

\section{Introduction}

Recent vision-language models (VLMs) have demonstrated increasingly strong general-purpose multimodal reasoning capabilities, yet these advances have not fully transferred to visual spatial planning~\citep{bai2025qwen3vltechnicalreport,bai2025qwen25vltechnicalreport,zhu2025internvl3exploringadvancedtraining,anthropic2025claudehaiku45,openai2024gpt4ocard}. Visual planning requires models to first recover task-relevant state structures from pixels and then reason over the recovered structures to generate executable actions~\citep{wu2024vspassessingdualchallenges,ivanitskiy2023configurablemaze,jin2023minibehavior,merler2026viplanbenchmarkvisualplanning,xu2026visualplanningletsthink}. This stepwise process distinguishes visual planning from symbolic planning. Symbolic inputs directly specify objects, constraints, and state transitions, while visual inputs require models to infer these elements before planning. This difference reveals a \emph{perception--reasoning modality gap} with two sequential bottlenecks: visual state recovery and plan generation. Perception errors in the first stage can affect subsequent planning, while the second stage may still fail even with a correctly recovered state.

\begin{figure}[!t]
\centering
\includegraphics[width=\columnwidth]{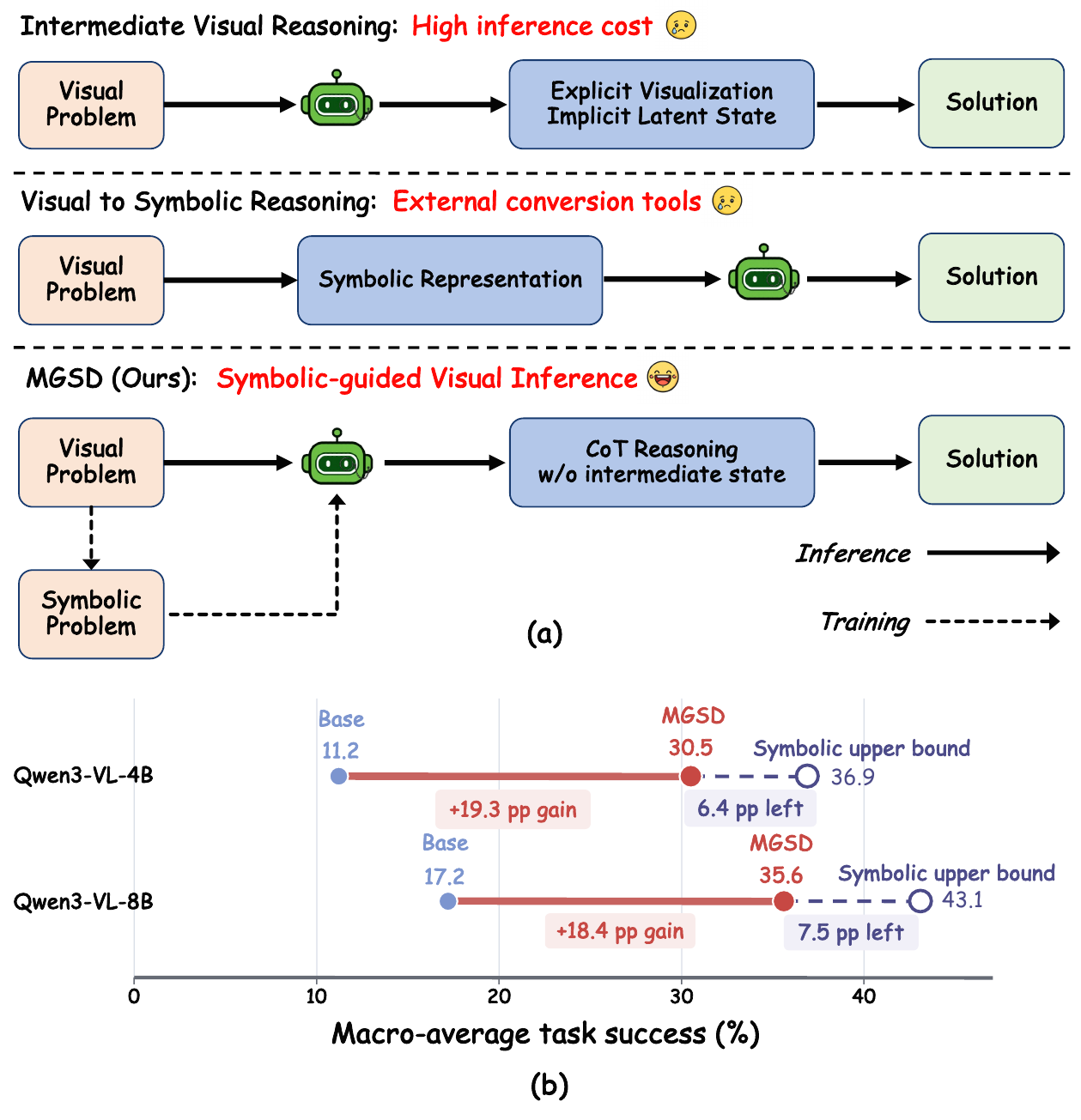}
\caption{
    \textbf{Overview and Effectiveness of \ours{}.}
    \textit{(a)} Comparison of visual reasoning paradigms. \ours{} uses symbolic supervision only during training and requires only visual inputs during inference.
    \textit{(b)} Comparison with the base models and corresponding symbolic-input upper bounds.
}
\label{fig:introduction}
\end{figure}

Existing training paradigms provide only partial solutions to this gap. Supervised fine-tuning (SFT) provides dense supervision by training models to reproduce fixed expert trajectories~\citep{xu2026visualplanningletsthink}. However, it offers no direct corrective guidance once generation deviates from the reference, and an incorrect action alone does not reveal whether the failure arises from visual state recovery, plan generation, or both~\citep{wu2024vspassessingdualchallenges,yang2025embodiedbench,merler2026viplanbenchmarkvisualplanning}. In contrast, reinforcement learning with verifiable rewards (RLVR) optimizes model-generated trajectories and has shown strong performance in multimodal reasoning~\citep{huang2026visionr1incentivizingreasoningcapability,shen2025vlmr1stablegeneralizabler1style}. However, standard RLVR typically assigns a single sequence-level reward based on the final answer, providing only coarse supervision for state recovery and planning decisions~\citep{miao2026seeingyouperceptionreasoningcoevolution,nguyen2026prpoparagraphlevelpolicyoptimization}. Together, these limitations reveal a supervision trade-off: SFT provides dense supervision on fixed trajectories, whereas RLVR trains on model-generated trajectories but relies on sparse sequence-level rewards. This motivates on-policy distillation (OPD), which trains on prefixes generated by the student while using a teacher to provide dense token-level supervision~\citep{tan2023gkdgeneralknowledgedistillation,zhao2026selfdistilledreasoneronpolicyselfdistillation,guo2026teachermovetemporalcoupling}. However, existing OPD generally assumes that the teacher and student receive the same input modality, limiting its ability to use symbolic states to guide a student that relies on visual observations.

To overcome this limitation, we propose \textbf{\ours{}}, a two-stage modality-gap-aware self-distillation framework for visual planning. Perception-oriented SFT first trains the visual student to recover task-relevant state structures from pixels. The student then generates rollout prefixes from visual inputs, while a text-only teacher conditions on these prefixes and the privileged symbolic context to provide token-level supervision. The student learns from the teacher by minimizing the reverse KL divergence between their next-token distributions at each prefix. As illustrated in Figure~\ref{fig:introduction}(a), symbolic information is used only during training, while inference remains entirely visual.

We evaluate \ours{} on visual planning benchmarks covering safe grid navigation~\citep{wu2024vspassessingdualchallenges}, topology-aware pathfinding~\citep{ivanitskiy2023configurablemaze}, and embodied object interaction~\citep{jin2023minibehavior}. As summarized in Figure~\ref{fig:introduction}(b), \ours{} improves the macro-average task success by 19.3 and 18.4 percentage points for the 4B and 8B backbones, respectively, narrowing the gaps to the corresponding symbolic-input upper bounds to 6.4 and 7.5 points. Ablation studies show that both training stages are important, while diagnostic analyses attribute the gains to improvements in visual state recovery and optimal-path reasoning. These results demonstrate that modality-gap-aware self-distillation improves both the perception of task-relevant states and the ability to plan over the recovered structures.

Our contributions are as follows:
\begin{itemize}

    \item We formulate visual spatial planning as a \textit{perception--reasoning modality gap} between visual and symbolic representations, with two sequential bottlenecks: recovering task-relevant states from pixels and generating valid plans from the recovered states.

    \item We propose \ours{}, a modality-gap-aware on-policy self-distillation framework for visual spatial planning that uses privileged symbolic states to supervise a visual student along its own rollout prefixes. It combines perception-oriented SFT with symbol-guided OPSD and requires only visual inputs during inference.
    
    \item Extensive experiments show that \ours{} substantially improves visual planning performance across different model scales and narrows the gap to symbolic-input upper bounds. Ablation and diagnostic results further demonstrate consistent improvements in both visual state recovery and optimal-path reasoning.

\end{itemize}

\begin{figure*}[t]
\centering
\includegraphics[width=\textwidth]{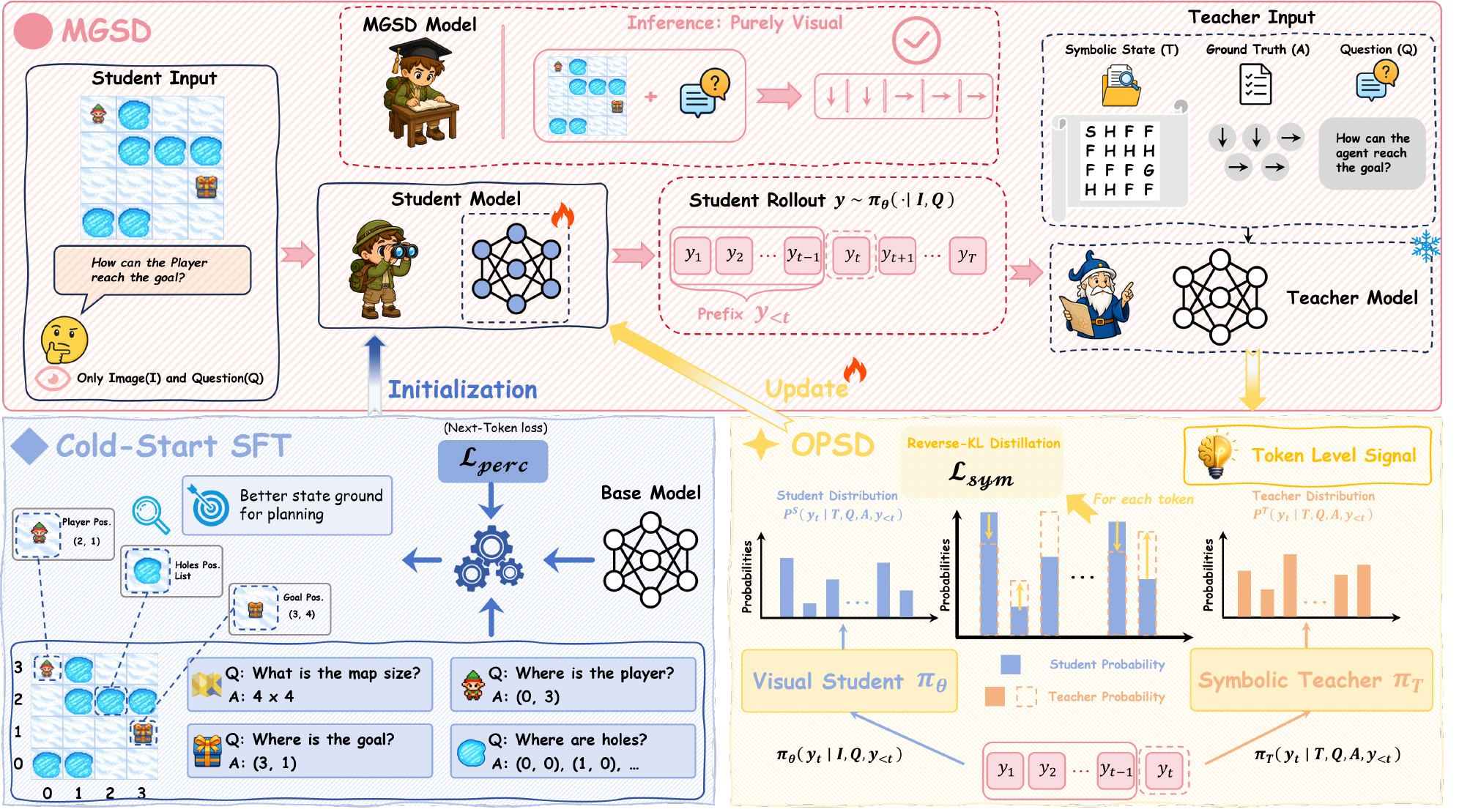}
\caption{
\textbf{Overview of the \ours{} framework.} Training consists of two stages to bridge the \emph{perception--reasoning modality gap}. \textbf{Bottom Left (Cold-Start SFT):} The base model is fine-tuned on structured perception tasks to reliably recover state variables (e.g., coordinates of the player, goal, and holes) from images. \textbf{Bottom Right (OPSD):} Symbol-guided on-policy self-distillation. The student generates reasoning rollouts from the image ($I$), while a privileged symbolic teacher conditions on the explicit symbolic state ($T$) and reference plan ($A$) to provide dense, token-level supervision on the student's prefix. \textbf{Top (Inference):} The symbolic teacher is discarded, and the trained student performs spatial planning purely from visual inputs.
}
\label{fig:overview}
\end{figure*}

\section{Related Work}
    
\paragraph{Visual Spatial Planning and the Modality Gap.}
Visual spatial planning requires VLMs to recover task-relevant states from pixels and reason over them to generate valid plans~\citep{wu2024vspassessingdualchallenges,ivanitskiy2023configurablemaze,jin2023minibehavior}. Fundamentally, such visual reasoning and state understanding greatly benefit from fine-grained visual grounding, such as parameter-efficient referring image segmentation and segmentation-based reward feedback~\citep{xu2023bridgingvisionlanguageencoders,huang2026samr1leveragingsamreward,huang2025denselyconnectedparameterefficienttuning}. To specifically tackle complex spatial planning, Existing approaches mainly follow two directions. One augments reasoning with intermediate representations, including generated visualizations, image sequences, and latent visual tokens~\citep{li2025imaginereasoningspacemultimodal,xu2026visualplanningletsthink,asadi2026mirageillusionvisualunderstanding,jin2026latentumunleashingpotentialinterleaved}. The other introduces an explicit visual-to-symbolic pipeline for formal planning~\citep{merler2026viplanbenchmarkvisualplanning,dang2025planningvisionlanguagemodelsuse,hao2026simulationrulesdualvlmframework}. These approaches generally add intermediate representations or conversion steps during inference. In contrast, our method uses symbolic states only as privileged supervision during training and performs inference directly from visual inputs.
    
\paragraph{On-Policy Distillation and Self-Distillation.}

Offline distillation trains the student on fixed sequences, creating a train--inference mismatch when it encounters unseen self-generated prefixes~\citep{tan2023gkdgeneralknowledgedistillation}. Representative methods address this issue through reverse-KL optimization and supervision on sequences sampled from the current student policy~\citep{jin2023minibehavior}, and recent efforts have further improved the stability of this process by investigating the temporal coupling of teacher updates~\citep{guo2026should}. Recent work extends this paradigm to reasoning and multimodal learning using privileged contexts, cross-modal teachers, additional visual evidence~\citep{zhao2026selfdistilledreasoneronpolicyselfdistillation,bousselham2026voldreasoningtransferllms,yuan2026visionopdlearningfinedetails,cai2026thinkingimagesinternalizingvisual}, or novel policy optimization strategies for autoregressive visual alignment~\citep{zhang2026kvpo}. In contrast, \ours{} uses explicit symbolic planning states as teacher-only supervision for an image-conditioned student along its own rollout prefixes. Symbolic information is restricted to training, while inference relies only on visual inputs.
    
\section{Method}
\label{sec:method}

\subsection{MGSD Framework Overview}
\label{sec:method-overview}

We propose \textbf{\ours{}}, a modality-gap-aware self-distillation framework for visual spatial planning, as illustrated in Figure~\ref{fig:overview}. Formally, each training instance is represented as $(I,Q,T,A)$, where $I$ denotes an image, $Q$ denotes a planning question, $T$ denotes the corresponding symbolic representation, and $A$ denotes a reference executable plan. The student is conditioned only on $(I,Q)$ and must recover the task state from pixels. The teacher has access to the privileged symbolic context $(T,Q,A)$ during training and can reason directly over explicit states. \ours{} uses this asymmetric setting to provide symbolic guidance to the visual student while preserving visual-only inference.

To exploit this asymmetric supervision, \ours{} adopts a two-stage training procedure that first establishes reliable visual grounding and then transfers planning behavior. The first stage performs \emph{perception-oriented supervised fine-tuning}. Instead of directly supervising long-horizon planning, this stage trains the student to recover task-relevant state information from pixels, providing a reliable basis for subsequent planning. The second stage performs \emph{symbol-guided on-policy self-distillation}. The student samples rollouts from its current policy using $(I,Q)$, while the teacher conditions on the privileged symbolic context and each student-generated prefix to provide token-level supervision.

\begin{algorithm}[t]
\small
\DontPrintSemicolon
\caption{\ours{}: Modality-Gap-Aware Self-Distillation}
\label{alg:mgsd}

\KwIn{
Base VLM $\pi_{\theta_0}$;
perception data $\mathcal{D}_{\mathrm{perc}}$;
planning data $\mathcal{D}=\{(I,Q,T,A)\}$
}
\KwOut{Visual planning model $\pi_{\theta^*}$}

$\pi_{\theta}\leftarrow\pi_{\theta_0}$\;

\vspace{0.15em}
\textbf{Stage 1: Perception-Oriented SFT}\;

\For{$\mathcal{B}_{\mathrm{perc}}\sim\mathcal{D}_{\mathrm{perc}}$}{
    $\theta\leftarrow\theta-\eta_{\mathrm{p}}
    \nabla_{\theta}\mathcal{L}_{\mathrm{perc}}
    (\theta;\mathcal{B}_{\mathrm{perc}})$
    \tcp*[r]{Eq.~\ref{eq:perc-loss}}
}

$\pi_{\mathrm{T}}\leftarrow
\operatorname{Freeze}(\pi_{\theta_0})$\;

\vspace{0.15em}
\textbf{Stage 2: Symbol-Guided OPSD}\;

\For{$\mathcal{B}
=\{(I_i,Q_i,T_i,A_i)\}_{i=1}^{B}\sim\mathcal{D}$}{

    \For{$i\leftarrow1$ \KwTo $B$}{
        $X_i\leftarrow(I_i,Q_i)$,
        $Z_i\leftarrow(T_i,Q_i,A_i)$\;

        $y_i\sim\pi_{\theta}(\cdot\mid X_i)$\;

        \For{$t\leftarrow1$ \KwTo $|y_i|$}{
            $p^{s}_{i,t}\leftarrow
            \pi_{\theta}(\cdot\mid X_i,y_{i,<t})$\;

            $p^{T}_{i,t}\leftarrow
            \pi_{\mathrm{T}}(\cdot\mid Z_i,y_{i,<t})$\;
        }
    }

    $\displaystyle
    \mathcal{L}_{\mathrm{sym}}
    \leftarrow
    \frac{1}{B}
    \sum_{i=1}^{B}
    \sum_{t=1}^{|y_i|}
    D_{\mathrm{KL}}
    \left(
    p^{s}_{i,t}
    \middle\|
    p^{T}_{i,t}
    \right)$
    \tcp*[r]{Eq.~\ref{eq:mgsd-loss}}

    $\theta\leftarrow\theta-\eta_{\mathrm{sym}}
    \nabla_{\theta}\mathcal{L}_{\mathrm{sym}}$\;
}

$\theta^*\leftarrow\theta$\;
\Return{$\pi_{\theta^*}$}\;

\end{algorithm}

\subsection{Perception-Oriented Supervised Fine-Tuning}
\label{sec:perception-sft}

Prior work shows that cold-start training can reduce the initial gap between student and teacher policies, thereby improving on-policy distillation~\citep{li2026rethinkingonpolicydistillationlarge,bousselham2026voldreasoningtransferllms}. We therefore introduce \emph{perception-oriented SFT} to reduce this early gap before OPSD. Instead of using complete trajectories as supervision, this stage trains the student to recover task-relevant states from pixels. The structured perception questions used for training are automatically generated from symbolic environment annotations. This stage improves visual grounding and provides a stronger initialization for subsequent training.

Let $\mathcal{D}_{\mathrm{perc}}$ denote the perception SFT dataset. Each example is a tuple $(I,Q_{\mathrm{p}},Y_{\mathrm{p}})$ containing an image, a perception question, and a structured target response. Given the question $X_{\mathrm{p}}=(I,Q_{\mathrm{p}})$, we minimize the standard autoregressive negative log-likelihood over the target response:
\begin{equation}
\label{eq:perc-loss}
\mathcal{L}_{\mathrm{perc}}(\theta)
=
-\mathbb{E}_{(I,Q_{\mathrm{p}},Y_{\mathrm{p}})\sim\mathcal{D}_{\mathrm{perc}}}
\left[
\sum_{t=1}^{|Y_{\mathrm{p}}|}
\log
\pi_{\theta}
\left(
Y_{\mathrm{p},t}
\,\middle|\,
X_{\mathrm{p}},Y_{\mathrm{p},<t}
\right)
\right].
\end{equation}

\subsection{Symbol-Guided On-Policy Self-Distillation}
\label{sec:opcd}

After SFT, we further optimize the student through \emph{symbol-guided OPSD}. For each training instance $(I,Q,T,A)$, the student uses only the visual observation $I$ and planning question $Q$, while the text-only teacher uses the privileged symbolic context $(T,Q,A)$. The student samples a trajectory from its current policy:
\begin{equation}
\label{eq:on-policy-rollout}
y=(y_1,\ldots,y_{|y|})
\sim
\pi_{\theta}(\cdot\mid I,Q).
\end{equation}

At each step $t$, both models compute their respective next-token distributions conditioned on the student-generated prefix $y_{<t}$. The student predicts from the visual context $(I,Q)$, whereas the teacher predicts from the symbolic context $(T,Q,A)$. At each step $t$, let $v\in\mathcal{V}$ denote a candidate token from the vocabulary. The student and teacher next-token probabilities are:
\begin{equation}
\label{eq:token-distributions}
\begin{aligned}
p^{s}_{t}(v)
&=
\pi_{\theta}(v\mid I,Q,y_{<t}),\\
p^{T}_{t}(v)
&=
\pi_{\mathrm{T}}(v\mid T,Q,A,y_{<t}).
\end{aligned}
\end{equation}

We optimize the student by minimizing the token-level reverse KL divergence between these distributions:
\begin{equation}
\label{eq:mgsd-loss}
\mathcal{L}_{\mathrm{sym}}(\theta)
=
\mathbb{E}_{y\sim\pi_{\theta}(\cdot\mid I,Q)}
\left[
\sum_{t=1}^{|y|}
D_{\mathrm{KL}}
\left(
p^{s}_{t}
\,\middle\|\,
p^{T}_{t}
\right)
\right],
\end{equation}
where
\begin{equation}
\label{eq:reverse-kl}
D_{\mathrm{KL}}
\left(
p^{s}_{t}
\middle\|
p^{T}_{t}
\right)
=
\sum_{v\in\mathcal{V}}
p^{s}_{t}(v)
\log
\frac{p^{s}_{t}(v)}
     {p^{T}_{t}(v)}.
\end{equation}
During self-distillation, the teacher remains frozen, and gradients update only the student. By matching the teacher distribution at student-generated prefixes, the objective provides dense supervision along the current student policy. Privileged symbolic information is restricted to training, and inference requires only $(I,Q)$.
    
\subsection{Training Procedure}
\label{sec:training-procedure}

Algorithm~\ref{alg:mgsd} summarizes the complete training procedure of \ours{}. We first optimize $\mathcal{L}_{\mathrm{perc}}$ on $\mathcal{D}_{\mathrm{perc}}$ and use the resulting parameters to initialize the student. During OPSD, the student samples trajectories from visual inputs, and the frozen teacher evaluates the same prefixes using the privileged symbolic context. We then update only the student by minimizing $\mathcal{L}_{\mathrm{sym}}$ in Eq.~\ref{eq:mgsd-loss}. After training, only the visual student is retained for inference.

\section{Experiments}
\label{sec:experiments}

\subsection{Experimental Setup}
\label{sec:exp-setup}
    
\begin{table*}[t]
\centering
\setlength{\tabcolsep}{4.0pt}
\resizebox{\textwidth}{!}{
\begin{tabular}{lcccccc c cccc c cc c c}
\toprule
\multirow{2}{*}{\textbf{Model}} 
& \multicolumn{6}{c}{\textbf{FrozenLake}} 
& \multirow{2}{*}{\textbf{Avg. F}} 
& \multicolumn{4}{c}{\textbf{Maze}} 
& \multirow{2}{*}{\textbf{Avg. M}} 
& \multicolumn{2}{c}{\textbf{MiniBehaviour}} 
& \multirow{2}{*}{\textbf{Avg. MB}} 
& \multirow{2}{*}{\textbf{Avg.}} \\
\cmidrule(lr){2-7} 
\cmidrule(lr){9-12} 
\cmidrule(lr){14-15}
& 3 & 4 & 5 & 6 & 7 & 8 
& 
& 3 & 4 & 5 & 6 
& 
& 5 & 6 
& 
& \\
\midrule
\rowcolor[HTML]{B7DDD4} 
\multicolumn{17}{c}{\textbf{Private Models }} \\
Claude-4.5-Haiku
& 49 & 41 & 29 & 17 & 10 & 6 & 25.3 
& 34.8 & 30.0 & 21.6 & 15.2 & 25.4 
& 12.3 & 11.1 & 11.7 
& 20.8 \\
GPT-4o
& 61 & 38 & 28 & 15 & 12 & 13 & 27.8 
& 38.0 & 30.4 & 20.4 & 16.8 & 26.4 
& 5.9 & 9.0 & 7.5 
& 20.6 \\
GPT-5
& 92 & 65 & 53 & 30 & 38 & 32 & 51.7 
& 77.2 & 44.4 & 26.4 & 16.0 & 41.0 
& 33.3 & 29.1 & 31.2 
& 41.3 \\
Gemini-2.5-Flash
& 92 & 91 & 74 & 58 & 37 & 33 & 64.2 
& 44.0 & 32.8 & 22.8 & 15.6 & 28.8 
& 27.0 & 22.1 & 24.5 
& 39.2 \\
Gemini-2.5-Pro
& 99 & 97 & 81 & 46 & 44 & 48 & 69.2 
& 41.2 & 21.6 & 13.6 & 8.8 & 21.3 
& 42.2 & 29.1 & 35.7 
& 42.1 \\
Gemini-3-Flash
& 100 & 100 & 99 & 96 & 84 & 97 & 96.0 
& 75.6 & 54.4 & 30.0 & 16.4 & 44.1 
& 65.7 & 56.3 & 61.0 
& 67.0 \\

\midrule
\rowcolor[HTML]{F1E1D5} 
\multicolumn{17}{c}{\textbf{Open-Source Models}} \\
LLaVA-OneVision-7B 
& 13 & 8 & 4 & 3 & 3 & 1 & 5.3 
& 7.2 & 4.0 & 5.6 & 4.4 & 5.3 
& 1.0 & 1.0 & 1.0 
& 3.9 \\
InternVL3-8B 
& 30 & 18 & 14 & 10 & 7 & 9 & 14.7 
& 10.0 & 7.6 & 7.6 & 4.0 & 7.3 
& 1.5 & 1.0 & 1.2 
& 7.7 \\
Qwen2.5-VL-3B-Instruct 
& 14 & 8 & 11 & 7 & 4 & 4 & 8.0 
& 10.0 & 5.6 & 5.6 & 5.2 & 6.6 
& 2.0 & 1.0 & 1.5 
& 5.4 \\
Qwen2.5-VL-7B-Instruct 
& 18 & 18 & 11 & 8 & 7 & 6 & 11.3 
& 11.2 & 6.0 & 10.8 & 3.6 & 7.9 
& 1.0 & 0.0 & 0.5 
& 6.6 \\
Qwen3-VL-4B-Instruct 
& 33 & 29 & 19 & 5 & 3 & 5 & 15.7 
& 24.4 & 12.0 & 5.2 & 4.8 & 11.6 
& 8.8 & 3.5 & 6.2 
& 11.2 \\
Qwen3-VL-8B-Instruct 
& 42 & 30 & 27 & 18 & 16 & 13 & 24.3
& 35.2 & 20.0 & 15.2 & 8.8 & 19.8 
& 8.3 & 7.0 & 7.7 
& 17.2 \\
Qwen3-VL-32B-Thinking
& 98 & 79 & 59 & 35 & 28 & 21 & 53.3 
& 54.4 & 34.0 & 19.2 & 17.6 & 31.3 
& 11.8 & 12.6 & 12.2 
& 32.3 \\
Qwen3-VL-235B-A22B-Instruct
& 77 & 56 & 44 & 20 & 20 & 18 & 39.2 
& 32.0 & 24.0 & 22.4 & 16.4 & 23.7 
& 13.2 & 15.6 & 14.4 
& 25.8 \\
Kimi-K2.5
& 93 & 77 & 75 & 54 & 48 & 51 & 66.3 
& 58.4 & 43.6 & 33.6 & 20.4 & 39.0 
& 16.2 & 16.6 & 16.4 
& 40.6 \\


\midrule
\rowcolor[HTML]{E9A36F} 
\multicolumn{17}{c}{\textbf{Our Models}} \\
\rowcolor[HTML]{DCEBFF} 
\textbf{\ours{}-4B} 
& \textbf{76} & \textbf{56} & \textbf{60} & \textbf{27} & \textbf{27} & \textbf{24} & \textbf{45.0} 
& \textbf{48.6} & \textbf{35.6} & \textbf{21.6} & \textbf{14.8} & \textbf{29.7} 
& \textbf{20.6} & \textbf{13.1} & \textbf{16.8} 
& \textbf{30.5} \\
\rowcolor[HTML]{F2F7FF} 
$\Delta$ vs. Qwen3-VL-4B 
& +43 & +27 & +41 & +22 & +24 & +19 & +29.3 
& +24.2 & +23.6 & +16.4 & +10.0 & +18.1 
& +11.8 & +9.6 & +10.6 
& +19.3 \\

\rowcolor[HTML]{DCEBFF} 
\textbf{\ours{}-8B} 
& \textbf{78} & \textbf{61} & \textbf{58} & \textbf{39} & \textbf{35} & \textbf{40} & \textbf{51.8} 
& \textbf{52.0} & \textbf{36.0} & \textbf{28.4} & \textbf{18.0} & \textbf{33.6} 
& \textbf{25.0} & \textbf{18.1} & \textbf{21.5} 
& \textbf{35.6} \\
\rowcolor[HTML]{F2F7FF} 
$\Delta$ vs. Qwen3-VL-8B 
& +36 & +31 & +31 & +21 & +19 & +27 & +27.5 
& +16.8 & +16.0 & +13.2 & +9.2 & +13.8 
& +16.7 & +11.1 & +13.8 
& +18.4 \\

\midrule
\rowcolor[HTML]{DCD3EA} 
\multicolumn{17}{c}{\textbf{Symbolic-Input Upper Bound}} \\

Symbolic-Input Qwen3-VL-4B
& 79 & 62 & 51 & 40 & 28 & 26 & 47.7
& 50.4 & 38.4 & 25.6 & 17.2 & 32.9
& 36.3 & 24.1 & 30.2
& 36.9 \\
\rowcolor[HTML]{F3F4F6}
$\Delta$ vs.  \ours{}-4B
& +3 & +6 & -9 & +13 & +1 & +2 & +2.7
& +1.8 & +2.8 & +4.0 & +2.4 & +3.2
& +15.7 & +11.0 & +13.4
& +6.4 \\

Symbolic-Input Qwen3-VL-8B
& 82 & 72 & 65 & 39 & 43 & 46 & 57.8
& 61.6/ & 46.0 & 31.6 & 29.6 & 42.2
& 31.9 & 26.6 & 29.2
& 43.1 \\
\rowcolor[HTML]{F3F4F6}
$\Delta$ vs. \ours{}-8B
& +4 & +11 & +7 & 0 & +8 & +6 & +6.0 
& +9.6 & +10.0 & +3.2 & +11.6 & +8.6 
& +6.9 & +8.5 & +7.7 
& +7.5 \\

\bottomrule
\end{tabular}
}
\caption{\textbf{Main results on visual planning benchmarks.}
We report task success accuracy across different difficulty levels.
Proprietary and open-source models are evaluated under prompt-only visual-input inference without task-specific training.
Symbolic-input models are reported as oracle upper bounds because they receive explicit state descriptions instead of images.
Task averages are computed within each environment, and Avg. denotes the macro average over environment-level averages.}
\label{tab:main-results}
\end{table*}
    

\paragraph{Benchmarks.}
We evaluate \ours{} on three visual planning environments: 
\textbf{FrozenLake}~\citep{wu2024vspassessingdualchallenges}, \textbf{Maze}~\citep{ivanitskiy2023configurablemaze}, and \textbf{MiniBehaviour}~\citep{jin2023minibehavior}. 
Following prior work, we evaluate executable task success and analyze failures related to state recovery and downstream planning~\citep{wu2024vspassessingdualchallenges,xu2026visualplanningletsthink}.
    
\paragraph{Training Data.}
We construct the data for both training stages from symbolic annotations across the three environments. Perception-oriented SFT uses 18K multimodal QA samples evenly distributed across the environments, while symbol-guided OPSD pairs each visual planning instance with its symbolic context and reference plan. We jointly train on mixed-task batches.
    
\paragraph{Baselines.}
We compare MGSD with both private and open-source VLMs under a unified evaluation protocol. The private baselines include Anthropic's Claude-4.5-Haiku~\citep{anthropic2025claudehaiku45}; OpenAI's GPT-4o and GPT-5~\citep{openai2024gpt4ocard,singh2026openaigpt5card}; and Google's Gemini-2.5-Flash, Gemini-2.5-Pro, and Gemini-3-Flash~\citep{comanici2025gemini25pushingfrontier,googledeepmind2025gemini3flash}.
These models are evaluated through their APIs with prompting only. 
The open-source baselines include LLaVA-OneVision-7B~\citep{li2024llavaonevisioneasyvisualtask}, InternVL3-8B~\citep{zhu2025internvl3exploringadvancedtraining}, Qwen2.5-VL~\citep{bai2025qwen25vltechnicalreport}, Qwen3-VL~\citep{bai2025qwen3vltechnicalreport}, and Kimi-K2.5 variants~\citep{kimiteam2026kimik25visualagentic}, all evaluated without task-specific fine-tuning. 
We also report symbolic-input upper bounds, where models receive explicit state representations rather than images. This setting bypasses visual state recovery and therefore serves as an oracle reference. It estimates the performance achievable when the underlying state is perceived without error.

\paragraph{Training Configuration.}
We train two variants of \ours{} using Qwen3-VL-4B-Instruct and Qwen3-VL-8B-Instruct as backbones. For each variant, the student is initialized from the pretrained checkpoint, and the same checkpoint is used to instantiate the teacher for OPSD. The teacher remains frozen throughout OPSD, while the student is updated. Training runs are conducted on 8 NVIDIA H200 GPUs.

\subsection{Main Results}
\label{sec:main-results}

\begin{figure*}[t]
\centering
\includegraphics[width=\textwidth]{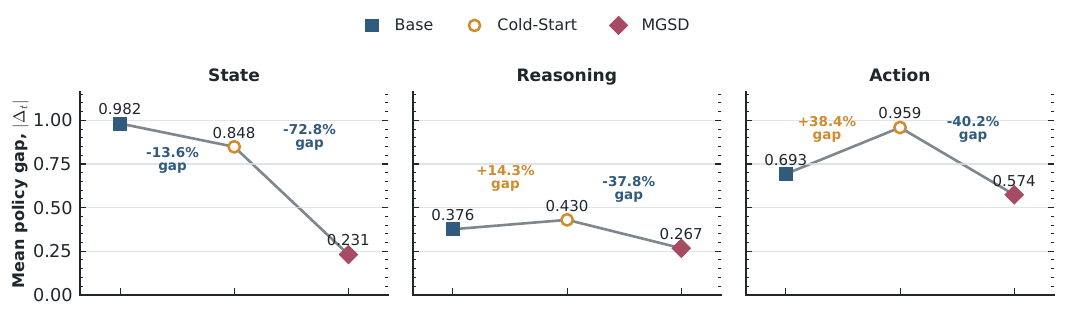}
\caption{
\textbf{Stage-wise teacher--student policy gap.} Points denote the macro-averaged absolute log-probability gap ($\vert{}\Delta_t\vert{}$) between the image-conditioned student and symbolic teacher. Lower values indicate better alignment. While Cold-Start mainly improves the State (perception) gap, \ours{} substantially aligns the policy across all three semantic regions. 
}
\label{fig:policy-gap}
\end{figure*}

Table~\ref{tab:main-results} reports the main task success results. \ours{} consistently improves both backbones across all three environments, increasing the macro-average score by 19.3\% and 18.4\% for the 4B and 8B models, respectively. Despite its compact scale, \ours{}-4B outperforms several larger open-source models and multiple private models. Moreover, relative to the base models, \ours{} reduces the gaps to the corresponding symbolic-input upper bounds by 75.1\% and 71.0\% for the 4B and 8B, respectively, demonstrating effective transfer of symbolic supervision to visual-only inference.

To examine how the gains emerge across different stages of generation, we measure the teacher--student policy gap over three regions of each student-generated response: state reconstruction (State), reasoning (Reasoning), and final action (Action). The gap is computed from the difference between the token probabilities assigned by the privileged teacher and the visual student.

As shown in Figure~\ref{fig:policy-gap}, while Cold-Start SFT reduces the macro-averaged State gap by 13.6\%, it inadvertently degrades downstream alignment, increasing the Reasoning and Action gaps by 14.3\% and 38.4\%. This confirms that improving visual state recovery alone does not guarantee text-aligned planning. In contrast, \ours{} distills the teacher's policy directly onto the student's visual prefixes, sharply reducing the State, Reasoning, and Action gaps to 0.231, 0.267, and 0.574 (relative reductions of 72.8\%, 37.8\%, and 40.2\% over Cold-Start). These dynamics demonstrate that \ours{} fundamentally aligns the student's reasoning and action policies with the privileged teacher.

\subsection{Ablation Study}
\label{sec:ablation}

\begin{table}[t]
\centering
\small
\setlength{\tabcolsep}{14pt}
\begin{tabular}{@{}l cc @{}}
\toprule
\textbf{Method Configuration} & \textbf{Acc.} & \textbf{Opt.} \\
\midrule

\multicolumn{3}{@{}l}{\textit{Standard Baselines}} \\
Base Model
    & 11.2 & 10.0 \\
Direct SFT
    & 18.2 & 15.9 \\
GRPO
    & 16.2 & 14.7 \\

\midrule
\multicolumn{3}{@{}l}{\textit{Framework Progression}} \\
Cold-Start SFT
    & 17.7 & 16.0 \\
\textbf{\ours{} (Full)}
    & \textbf{30.5} & \textbf{26.6} \\

\midrule
\multicolumn{3}{@{}l}{\textit{MGSD Ablations}} \\
\quad w/o Cold Start
    & 16.8 & 14.7 \\
\quad w/o Reference Plan
    & 21.3 & 18.3 \\
\quad w/o Symbolic State
    & 11.2 & 9.8 \\
\quad w/ EMA Teacher
    & 19.5 & 17.2 \\

\bottomrule
\end{tabular}
\caption{
\textbf{Ablation study of \ours{}.} 
Acc. denotes task success accuracy, and Opt. denotes optimal-path accuracy.
We compare training baselines, the contributions of the two training stages, and different sources of teacher-side supervision.
}
\label{tab:ablation}
\end{table}
    
Table~\ref{tab:ablation} reports the ablation results of \ours{} using Qwen3-VL-4B-Instruct as the backbone. We compare traditional training baselines, examine the contribution of each training stage, and evaluate the main design choices of \ours{}.

\paragraph{Baselines and Stage-wise Contributions.}
Traditional training methods provide only limited gains on visual planning. Direct SFT, which trains the model on reference plans using standard teacher forcing, achieves an accuracy of 18.2\%, while GRPO reaches 16.2\%, suggesting that neither imitation learning nor outcome-level reinforcement learning is sufficient for this task. Perception-oriented SFT alone obtains 17.7\%, providing the student with basic visual state grounding. Adding symbol-guided OPSD substantially improves the accuracy to 30.5\%. The cold-start stage provides the necessary visual grounding, while symbol-guided OPSD delivers the main improvement by transferring planning knowledge from the symbolic teacher to the student's own rollouts.

\paragraph{Design Choices of \ours{}.}
Removing the cold-start stage (\texttt{w/o Cold Start}) lowers the accuracy to 16.8\%, showing that the student needs reliable visual state understanding before learning planning behavior. Excluding the reference action sequence from the teacher context (\texttt{w/o Reference Plan}) reduces the accuracy to 21.3\%, because the teacher no longer has access to the exact target plan and therefore provides less precise guidance. Crucially, depriving the teacher of the privileged symbolic state input (\texttt{w/o Symbolic State})---forcing it to rely on the same visual input as the student alongside the reference plan---plummets the accuracy to 11.2\%. This severe drop back to base-model performance demonstrates that the explicit symbolic state is essential; without it, the teacher suffers from the same visual perception bottlenecks and fails to provide high-quality, dense reasoning supervision. Finally, replacing the frozen teacher with an exponential moving average teacher (\texttt{w/ EMA Teacher}) decreases the accuracy to 19.5\%. This result suggests that a fixed symbolic teacher offers a more stable and consistent distillation target than a teacher that evolves together with the student.

\subsection{Diagnostic Analysis: Decoupling Perception and Planning}
\label{sec:diagnostic}

\begin{figure}[t]
    \centering
    \includegraphics[width=\columnwidth]{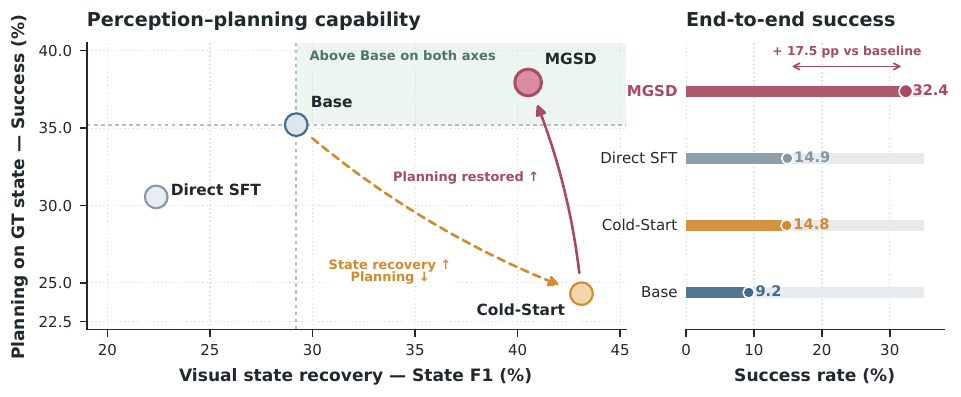}
\caption{
\textbf{Diagnostic decomposition of perception and planning.}
\textit{Left:} Visual state recovery measured by State F1 and planning performance measured on ground-truth symbolic states. 
\textit{Right:} End-to-end task success on the held-out diagnostic split. 
Cold-Start SFT improves state recovery but reduces planning performance, while \ours{} achieves a stronger balance between the two capabilities.
}
\label{fig:diagnostic_map}
\end{figure}

\begin{figure*}[t]
\centering
\includegraphics[width=\textwidth]{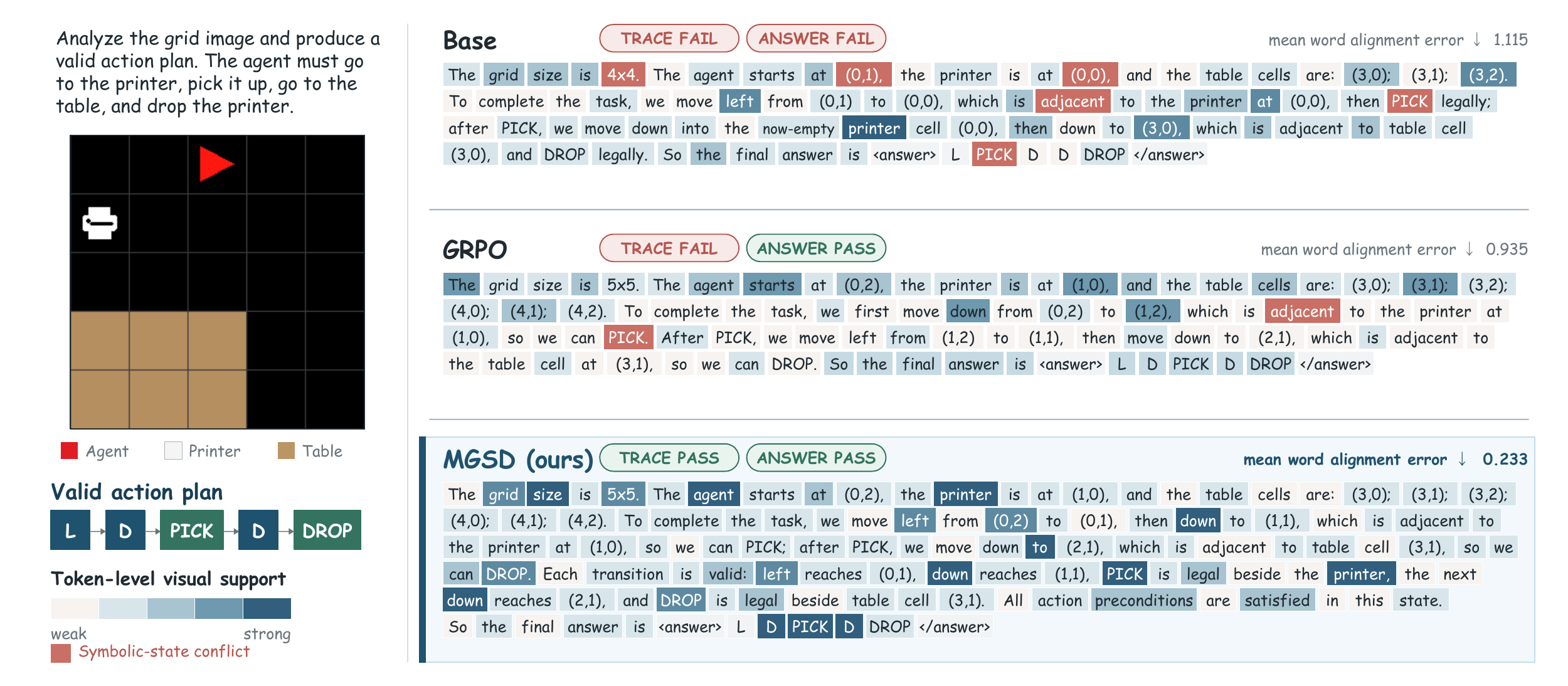}
\caption{
\textbf{Qualitative analysis of visual grounding and symbolic consistency.}
Blue intensity indicates token-level visual support estimated through counterfactual image interventions, while red highlights mark conflicts with the symbolic environment state or action rules. Mean word alignment error measures deviation from the reference support signal, with lower values indicating better alignment. In this example, \ours{} produces both a state-consistent reasoning trace and a valid action plan.
}
\label{fig:figure2}
\end{figure*}
    
End-to-end task success combines two capabilities: recovering the task state from visual inputs and generating a valid plan over the recovered state. We separate these capabilities using three metrics: \textbf{State F1}, which measures visual state recovery; \textbf{Plan on GT}, which measures planning performance given ground-truth symbolic states; and \textbf{E2E Acc.}, which measures standard visual-to-action performance. These results are evaluated on a held-out diagnostic split, so the E2E values differ from those in Table~\ref{tab:main-results}.

As shown in Figure~\ref{fig:diagnostic_map}, Cold-Start SFT substantially improves State F1 from 29.2\% to 43.1\%, but reduces Plan on GT from 35.2\% to 24.3\%. This results show that improved visual state recovery does not necessarily translate into better planning. Moreover, the base model achieves only 35.2\% when directly given ground-truth symbolic states, indicating that planning remains a separate source of error even when perception is removed.

\ours{} improves the balance between these two capabilities. It retains most of the state-recovery gain obtained by Cold-Start SFT while increasing Plan on GT to 37.9\%. As a result, \ours{} achieves the highest end-to-end success rate among the compared methods on the diagnostic split, reaching 32.4\%. These results indicate that symbol-guided OPSD improves planning performance without substantially sacrificing visual state recovery.

\subsection{Qualitative Analysis of Visual Grounding}
\label{sec:qualitative}

End-to-end accuracy does not reveal whether a correct answer is supported by valid visual grounding. Following prior counterfactual visual-dependency analysis~\citep{liu2026visualadvantageonpolicy}, we analyze a MiniBehaviour example using counterfactual image interventions. For each fixed model-generated rollout, we keep the prompt and generated prefix unchanged and recompute token probabilities after task-relevant entities are removed or moved. The resulting log-probability change measures token-level visual support: larger positive values indicate stronger support from the original image, whereas negative values indicate that the image provides weaker support than the counterfactual inputs. In Figure~\ref{fig:figure2}, blue intensity visualizes this support. Symbolic conflicts are detected separately by verifying generated state descriptions and actions against the environment state and transition rules, and are highlighted in red.

We additionally report the mean word alignment error, which measures the deviation between the visual-support signal of each model and a reference signal derived from privileged symbolic supervision. Subword scores are first aggregated within each word, and lower values indicate closer alignment with the reference signal. 

As shown in Figure~\ref{fig:figure2}, the Base model incorrectly recovers the environment state and consequently produces an invalid reasoning trace and final answer. GRPO generates a valid final action sequence, but its intermediate reasoning remains inconsistent with the environment dynamics. In contrast, \ours{} achieves the lowest alignment error and produces both a state-consistent reasoning trace and an executable action plan. This example illustrates that \ours{} improves the consistency among visual evidence, intermediate reasoning, and final action generation. By mitigating hallucinated reasoning, such alignment establishes a critical foundation for trustworthy and interpretable visual planning.

\section{Conclusion}
In this work, we introduced \ours{}, a two-stage modality-gap-aware self-distillation framework designed for visual spatial planning. By first aligning the visual student with planning-relevant state structures and subsequently distilling symbolic planning behavior onto student-generated rollouts, \ours{} effectively bridges the gap between image-conditioned perception and explicit symbolic planning. Extensive experiments across FrozenLake, Maze, and MiniBehaviour demonstrate substantial gains over base VLMs, significantly narrowing the performance gap relative to symbolic-input upper bounds. Furthermore, our diagnostic analyses confirm that \ours{} simultaneously improves visual state recovery and optimal-path reasoning, proving that the model is strengthened in both what it perceives and how it plans over the inferred state. Ultimately, our findings establish privileged symbolic supervision as a highly practical paradigm for advancing visual planning while maintaining a strictly visual inference pipeline.

\section*{Limitations}
    
\ours{} relies on paired visual and symbolic training data, where symbolic states and reference plans provide privileged supervision. This setting is natural for simulator-based environments, but may be harder to obtain in open-world scenes. Our evaluation focuses on structured tasks with discrete action spaces, while broader and less structured settings remain outside the current scope. Finally, although \ours{} improves visual state recovery and optimal-path reasoning, it does not guarantee faithful intermediate reasoning under distributional uncertainty. Extending the method to noisy or automatically extracted symbolic states and stronger intermediate-state verification remains future work.

\bibliography{preprint}

\clearpage

\appendix
\setcounter{secnumdepth}{1}

\section*{Supplementary Material}

\section{Training Details}
\label{app:training}

We train \ours{} in two stages. First, perception-oriented SFT adapts the model with LoRA on structured multimodal QA data derived from symbolic environment annotations; the resulting adapter is merged into the base model to initialize the next stage. Second, symbol-guided on-policy self-distillation trains the visual student with rollouts sampled from image-conditioned prompts, while a frozen text-only teacher uses symbolic contexts and reference action plans to provide token-level distillation signals. Table~\ref{tab:sft-training-details} and Table~\ref{tab:opsd-training-details} summarize the main hyperparameters for the two stages.

\begin{table}[t]
\centering

\small
\setlength{\tabcolsep}{5pt}
\renewcommand{\arraystretch}{1.05}
\begin{tabular}{@{}ll@{}}
\toprule
\textbf{Setting} & \textbf{Value} \\
\midrule
Backbone & Qwen3-VL-4B \\
Initialization & Base model \\
Training tasks & 3 tasks \\
Training samples & 18K \\
Validation samples & 2\% held-out \\
Tuning method & LoRA \\
Precision & bfloat16 \\
Epochs & 3 \\
Learning rate & $1\times10^{-4}$ \\
Scheduler & cosine \\
Warmup ratio & 0.1 \\
LoRA rank / alpha & 8 / 16 \\
LoRA dropout & 0.0 \\
Max sequence length & 2,048 \\
Validation frequency & every 100 steps \\
Checkpoint frequency & every 200 steps \\
\bottomrule
\end{tabular}
\caption{SFT training hyperparameters.}
\label{tab:sft-training-details}
\end{table}

\begin{table}[t]
\centering

\small
\setlength{\tabcolsep}{5pt}
\renewcommand{\arraystretch}{1.05}
\begin{tabular}{@{}ll@{}}
\toprule
\textbf{Setting} & \textbf{Value} \\
\midrule
Backbone & Qwen3-VL-4B \\
Initialization & SFT-merged model \\
Training tasks & 3 tasks \\
Training samples & 18K \\
Validation samples & 1.2K \\
Tuning method & On-policy distillation \\
Teacher & frozen text-only teacher \\
Precision & bfloat16 \\
Epochs & 3 \\
Max prompt length & 5,120 \\
Max response length & 2,048 \\
Rollouts per prompt & 1 \\
Rollout batch size & 32 \\
Actor global batch size & 32 \\
Validation frequency & every 20 steps \\
Checkpoint frequency & every 20 steps \\
Number of GPUs & 8 \\
\bottomrule
\end{tabular}
\caption{OPSD training hyperparameters.}
\label{tab:opsd-training-details}
\end{table}

\section{Data Construction Details}
\label{app:data_construction}

To systematically address the perception--reasoning modality gap, we construct the FrozenLake, Maze, and MiniBehaviour datasets by strictly pairing raw visual observations with corresponding symbolic states. This design guarantees that the privileged teacher supervision is grounded in verified graph-search solutions rather than heuristic rollouts or noisy model-generated text.

\subsection{Generation Principles}

Our procedural generation pipeline enforces four invariants to ensure label consistency and data quality:

\begin{enumerate}
    \item \textbf{Reproducibility:} All environment instances are procedurally generated using deterministic random seeds, uniquely defined by task level, difficulty bucket, and sampling index. 
    \item \textbf{Guaranteed Solvability:} Candidate states undergo rigorous graph-based validation. Solutions are computed via exact shortest-path algorithms; any candidate lacking a complete, legal solution is discarded.
    \item \textbf{Action Feasibility:} Generated trajectories are strictly verified against environment transition dynamics. Move actions must be locally valid, interactions must satisfy explicit state preconditions, and terminal states must strictly align with task success criteria.
    \item \textbf{Balanced Complexity:} To prevent distribution skew toward trivial or overly complex paths, we apply rejection sampling across predefined difficulty buckets. Intra-level deduplication (via spatial layout and key-state hashing) further maximizes dataset diversity.
\end{enumerate}

\subsection{Environment-Specific Construction}

\begin{itemize}
    \item \textbf{FrozenLake:} Formulated as an $N \times N$ grid navigation task. Obstacle density is controlled by modulating the frozen-cell sampling probability. We extract a graph of safe traversable cells and apply Breadth-First Search (BFS) to find the shortest path from start to goal, ensuring the reference trajectory strictly avoids all hazard tiles.
    \item \textbf{Maze:} Constructed using a ``perfect maze'' topology via randomized Depth-First Search (DFS). Perfect mazes are fully connected and acyclic, guaranteeing a unique simple path between any two coordinates. This topological constraint is vital for reasoning supervision, as it entirely eliminates optimal-path ambiguity. Difficulty is bucketed by the minimum path length.
    \item \textbf{MiniBehaviour:} Modeled as a two-stage embodied interaction task that requires evaluating dynamic reachability. Prior to the \texttt{PICK} action, both the target object (printer) and the destination region (table) act as non-traversable obstacles. Executing a valid \texttt{PICK} removes the object and updates the obstacle mask. The generator runs BFS independently for the start-to-printer and printer-to-table sub-tasks, ensuring sequential adjacency constraints are satisfied before stitching the actions into a unified trajectory.
\end{itemize}

\subsection{Dataset Statistics and Distribution}

\begin{figure*}[t]
\centering
\includegraphics[width=\textwidth]{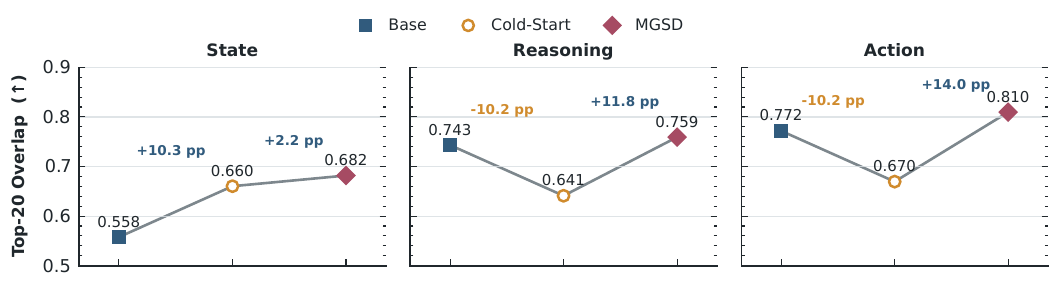}
\caption{
Stage-wise teacher--student Top-20 candidate overlap across State, Reasoning, and Action tokens. 
Values denote task-macro means, and labels between adjacent stages report percentage-point changes (higher is better). Marker shape and colour identify Base, Cold-Start, and \ours{}.
}
\label{fig:top20_overlap}
\end{figure*}

To ensure that the visual environments encountered during cold-start perception alignment match the complexity of those used for reasoning distillation, both the SFT and OPSD stages sample from the identical underlying distribution of 18,000 unique environment configurations (rendered as $256 \times 256$ RGB images):

\begin{itemize}
    \item \textbf{FrozenLake (6,000 instances):} Uniformly distributed across map levels 3 through 8 (1,000 per level). Each level is evenly stratified across five frozen-cell probability buckets (200 per bucket).
    \item \textbf{Maze (6,000 instances):} Uniformly distributed across map levels 3 through 6 (1,500 per level). Each level is evenly stratified across easy, medium, and hard path-length buckets (500 per bucket).
    \item \textbf{MiniBehaviour (6,000 instances):} Split equally between map levels 5 and 6 (3,000 per level). Each level is evenly stratified across easy, medium, and hard difficulty buckets (1,000 per bucket).
\end{itemize}
    
\subsection{SFT Target Formulation}
    
Rather than sampling separate states, the cold-start SFT dataset deterministically converts the verified symbolic annotations of the 18,000 training instances into structured QA pairs. Because these regression targets are algorithmically extracted from the underlying state engine, they provide noise-free perception grounding.
    
\begin{itemize}
    \item \textbf{FrozenLake:} Requires extracting grid size, player coordinates, goal coordinates, and an exhaustive list of hazardous hole coordinates.
    \item \textbf{Maze:} Requires parsing fine-grained topological structure by generating a complete open-direction reachability table for every valid cell based on the visual walls.
    \item \textbf{MiniBehaviour:} Requires grounding object coordinates, occupied regions, dynamically calculated legal adjacency sets, and current binary interaction affordances (e.g., verifying if \texttt{PICK}/\texttt{DROP} is legally permitted).
\end{itemize}
    
\subsection{Held-out Validation}
    
The validation set comprises 1,200 instances (600 FrozenLake, 400 Maze, 200 MiniBehaviour). These are generated using identical dynamics and difficulty criteria but are strictly isolated from the training corpus to provide a robust evaluation of out-of-distribution reasoning and perception.

\section{Detailed Analysis of Token-Region Policy Gap}
\label{app:token-gap}

This section provides the comprehensive experimental protocol, metric definitions, and granular task-level diagnostic results supporting the policy gap analysis presented in the corresponding section of the main paper.

\subsection{Experimental Protocol}
To rigorously evaluate how different training stages affect the model's internal planning behavior, we evaluated 1,200 procedurally generated held-out validation cases (600 FrozenLake, 400 Maze, 200 MiniBehaviour). For each case, the image-conditioned visual student performs greedy autoregressive decoding to generate a complete rollout sequence. 

Simultaneously, we evaluate this identical generated sequence using a fixed text-only teacher via teacher forcing. The student receives the raw visual image and task instructions as its context ($c_S$). The teacher, acting as a privileged oracle, receives a fully observable symbolic text context ($c_T$) that explicitly details the grid layout, object locations, constraints, and the reference trajectory. This asymmetric evaluation allows us to measure how closely the student's visual policy aligns with the teacher's symbolic policy on the student's own generated path.

\subsection{Token Region Definitions}
To localize the policy mismatch, we automatically parse the student-generated responses into three sequential semantic regions based on character offsets and task-specific regular expressions:
\begin{itemize}
    \item \textbf{State (Perception):} Tokens from the start of the response up to the first explicit reasoning boundary. This captures the model's visual state recovery (e.g., extracting map sizes, object coordinates, or wall topologies).
    \item \textbf{Reasoning (Planning):} Tokens from the start of the reasoning process up to the final action block. This captures the agent's logic in connecting the perceived state to an executable plan.
    \item \textbf{Action (Execution):} The precise sequence of executable steps enclosed within the \texttt{<answer>...</answer>} tags. The formatting tags themselves are excluded from the metric to prevent superficial formatting alignment from skewing the underlying decision policy gap.
\end{itemize}

\subsection{Evaluation Metrics}
We quantify the teacher--student discrepancy at the token level using two complementary metrics:

\paragraph{Absolute Log-Probability Gap.} 
For every selected rollout token $y_t$, we calculate the absolute difference in log-probabilities assigned by the teacher and the student:
$$|\Delta_t| = \left| \log \pi_T(y_t \mid c_T, y_{<t}) - \log \pi_S(y_t \mid c_S, y_{<t}) \right|$$
A lower value indicates that the teacher and student exhibit similar confidence for the selected token.

\paragraph{Top-$k$ Candidate Overlap.} 
To verify if the student and teacher are sampling from the same underlying token support regions, we calculate the candidate-support overlap~\citep{li2026rethinkingonpolicydistillationlarge}:
$$O_t(k) = \frac{|\operatorname{TopK}_S(t) \cap \operatorname{TopK}_T(t)|}{k}, \quad k \in \{5, 10, 20, 50\}$$
Values closer to 1 indicate that the student and teacher share highly consistent high-probability candidate sets, whereas values closer to 0 suggest severe distribution mismatch.

\subsection{Detailed Results and Analysis}

\paragraph{Stage-wise Trajectory (Macro Average).}
As summarized in the policy-gap figure in the main paper, the Base model struggles uniformly, showing significant absolute log-probability gaps for State (0.982), Reasoning (0.376), and Action (0.693). The Cold-Start perception SFT successfully reduces the State gap to 0.848 (a 13.6\% reduction), aligning with its goal of improving visual grounding. However, this perceptual adaptation inadvertently harms downstream alignment, pushing the Reasoning gap up to 0.430 (+14.3\%) and the Action gap up to 0.959 (+38.4\%). 

\ours{} effectively resolves this bottleneck. By distilling reasoning behavior directly onto the student's visual prefixes, \ours{} drops the State, Reasoning, and Action gaps to 0.231, 0.267, and 0.574, respectively. This corresponds to relative reductions of 72.8\%, 37.8\%, and 40.2\% over the Cold-Start stage, proving that the student successfully internalizes the teacher's planning distributions.

\paragraph{Top-20 Candidate Overlap.}
To further validate the alignment of the underlying token support regions, Figure~\ref{fig:top20_overlap} illustrates the Top-20 candidate overlap ($O_t@20$). Cold-Start increases the candidate overlap over the Base model in the State region (+10.2 percentage points, from 0.558 to 0.660), but severely reduces it in Reasoning and Action (-10.2 pp and -10.2 pp, respectively). \ours{} subsequently increases the overlap over Cold-Start by +2.2 pp (State), +11.8 pp (Reasoning), and +14.0 pp (Action), reaching 0.682, 0.759, and 0.810 respectively. These displayed trajectories separate the perceptual benefit of Cold-Start from the broader teacher--student candidate-support alignment obtained after \ours{}, demonstrating the largest recovery in Action tokens.

\paragraph{Task-Level Consistency.}

\begin{figure}[!t]
\centering
\includegraphics[width=\columnwidth]{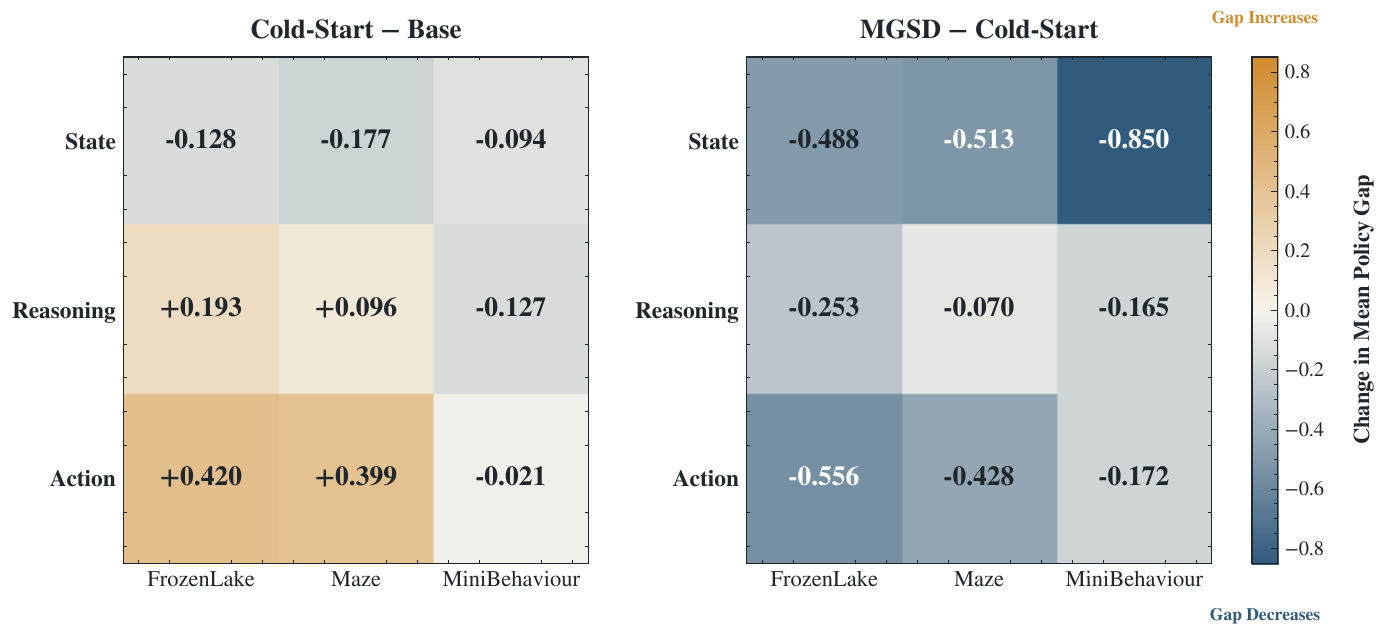}
\caption{
Task-level consistency of policy-gap changes. 
Each cell reports the change in the mean absolute selected-token log-probability gap between adjacent training stages for a specific VSP task and output region. Negative values (blue) indicate reduced teacher--student discrepancy, signifying an improvement in policy alignment. Positive values (gold) indicate an increased discrepancy. Cold-Start successfully reduces the State gap but introduces heterogeneous downstream degradation, whereas \ours{} uniformly reduces the policy gap across all nine task-by-region cells.
}
\label{fig:task_matrices}
\end{figure}

To ensure these macro-level gains are not dominated by outliers, Figure~\ref{fig:task_matrices} provides a granular breakdown of the mean policy gap changes across individual environments. 

During the Cold-Start stage, the State gap consistently decreases across all tasks: FrozenLake (-0.128), Maze (-0.177), and MiniBehaviour (-0.094). However, this purely perceptual alignment severely disrupts downstream sequence generation, drastically increasing the Action gap in FrozenLake (+0.420) and Maze (+0.399), as well as the Reasoning gap in both environments (+0.193 and +0.096, respectively).

The \ours{} stage demonstrates robust, universal improvement, with gap changes being strictly negative across all nine task-by-region cells. \ours{} substantially refines the State gap by -0.488 in FrozenLake, -0.513 in Maze, and -0.850 in MiniBehaviour. More crucially, it rectifies the downstream reasoning and action bottlenecks, lowering the Action gap by -0.556 in FrozenLake, -0.428 in Maze, and -0.172 in MiniBehaviour. These fine-grained results reinforce our conclusion that \ours{} consistently bridges the perception--reasoning modality gap across distinct spatial environments.

\section{Diagnostic Experiment Details}

\begin{table*}[t]
\centering
\small
\setlength{\tabcolsep}{3pt}
\begin{tabular}{@{}lccc@{\qquad}lccc@{}}
\toprule
\multicolumn{4}{c}{\textbf{FrozenLake}} &
\multicolumn{4}{c}{\textbf{Maze}} \\
\cmidrule(r){1-4}\cmidrule(l){5-8}
\textbf{Model} & State F1 & Plan on GT & E2E Acc. &
\textbf{Model} & State F1 & Plan on GT & E2E Acc. \\
\midrule
Base & 37.4 & 44.7 & 11.2 &
Base & 26.6 & 40.5 & 14.0 \\
Direct SFT & 29.6 & 38.7 & 18.5 &
Direct SFT & 13.0 & 36.5 & 17.2 \\
Cold-Start SFT & \textbf{41.0} & 32.2 & 20.7 &
Cold-Start SFT & 34.0 & 31.2 & 15.8 \\
\ours{} (Ours) & 40.9 & \textbf{42.3} & \textbf{39.3} &
\ours{} (Ours) & \textbf{43.2} & \textbf{44.5} & \textbf{35.8} \\
\midrule
Base + GT State & 100.0 & 44.7 & 44.7 &
Base + GT State & 100.0 & 40.5 & 40.5 \\
\bottomrule
\end{tabular}
\caption{Detailed diagnostic results on FrozenLake and Maze. State F1 measures visual state recovery, Plan on GT measures planning from ground-truth symbolic states, and E2E Acc. measures end-to-end task accuracy. Values are percentages.}
\label{tab:diagnostic-frozenlake-maze}
\end{table*}

\begin{table*}[t]
\centering
\small
\setlength{\tabcolsep}{3pt}
\begin{tabular}{@{}lccc@{\qquad}lccc@{}}
\toprule
\multicolumn{4}{c}{\textbf{MiniBehaviour}} &
\multicolumn{4}{c}{\textbf{Macro Average}} \\
\cmidrule(r){1-4}\cmidrule(l){5-8}
\textbf{Model} & State F1 & Plan on GT & E2E Acc. &
\textbf{Model} & State F1 & Plan on GT & E2E Acc. \\
\midrule
Base & 23.7 & 20.5 & 2.5 &
Base & 29.2 & 35.2 & 9.2 \\
Direct SFT & 24.6 & 16.5 & 9.0 &
Direct SFT & 22.4 & 30.6 & 14.9 \\
Cold-Start SFT & \textbf{54.4} & 9.5 & 8.0 &
Cold-Start SFT & \textbf{43.1} & 24.3 & 14.8 \\
\ours{} (Ours) & 37.4 & \textbf{27.0} & \textbf{22.0} &
\ours{} (Ours) & 40.5 & \textbf{37.9} & \textbf{32.4} \\
\midrule
Base + GT State & 100.0 & 20.5 & 20.5 &
Base + GT State & 100.0 & 35.2 & 35.2 \\
\bottomrule
\end{tabular}
\caption{Detailed diagnostic results on MiniBehaviour and the task-level macro average. The Base + GT State row is an upper-bound sanity check. Values are percentages.}
\label{tab:diagnostic-minibehaviour-macro}
\end{table*}
\label{app:diagnostic-details}
This section provides the detailed setup, metric definitions, and task-level breakdowns for the causal decomposition diagnostic discussed in the diagnostic section of the main paper. 

\paragraph{Dataset Discrepancy.} The End-to-End (E2E) accuracies reported here differ from the main-results table in the paper. By design, isolating pure reasoning (\textit{Plan on GT}) requires perfectly paired ground-truth symbolic states, which standard open benchmarks lack. Consequently, these diagnostic experiments exclusively evaluate our 1,200 procedurally generated validation samples (600 FrozenLake, 400 Maze, 200 MiniBehaviour) to ensure flawless perception-reasoning alignment.

\subsection{Experimental Design and Metrics}
To diagnose whether visual spatial planning failures stem from perception errors or reasoning bottlenecks, we decouple the end-to-end evaluation into three complementary metrics:
\begin{itemize}
    \item \textbf{State F1:} Evaluates the model's ability to visually recover the task's structural state. For FrozenLake, this includes grid size, agent/goal coordinates, and the exact hole set. For Maze, it requires the precise open-direction table (topology) for every cell. For MiniBehaviour, it requires object locations, table cells, and valid \texttt{PICK}/\texttt{DROP} affordances.
    \item \textbf{Plan on GT:} The model is provided strictly with the text-based ground-truth symbolic state (no images) and must output a plan. This isolates the model's pure reasoning capabilities by removing visual perception errors.
    \item \textbf{E2E Acc.:} The standard end-to-end evaluation where the model receives an image and directly outputs an executable action plan.
\end{itemize}

\subsection{Detailed Task-Level Results}

Tables~\ref{tab:diagnostic-frozenlake-maze} and~\ref{tab:diagnostic-minibehaviour-macro} present the full causal decomposition across all three tasks alongside the macro average. The task-level breakdowns reinforce our main claims regarding the perception--reasoning modality gap.

\textbf{Perception Gap:} All tested base VLMs struggle to recover perfectly actionable symbolic states from images. In Maze, where fine-grained topological extraction is required, visual state recovery is particularly challenging. However, both Cold-Start SFT and \ours{} drastically improve State F1 across all tasks compared to the Base and Direct SFT models, proving that dedicated perception alignment is crucial.

\textbf{Reasoning Gap:} As established in the main text, providing the Base model with ground-truth symbolic text (Base + GT State) only yields a 35.2\% macro success rate. This proves that even with perfect perception, VLMs lack the intrinsic reasoning to consistently generate valid spatial plans. \ours{} achieves the highest Plan on GT performance among the trained models, specifically reaching 42.3\% on FrozenLake, 44.5\% on Maze, and 27.0\% on MiniBehaviour. 

Crucially, the task-level data illustrates why solely optimizing for perception is insufficient. In MiniBehaviour, Cold-Start SFT alone achieves the highest perception accuracy (54.4\% State F1) but fails severely at reasoning (9.5\% Plan on GT), resulting in a poor E2E accuracy (8.0\%). \ours{} demonstrates the best synergy: by distilling reasoning behaviors directly onto the student's visual prefixes, it balances high-quality visual state recovery with robust reasoning, achieving state-of-the-art End-to-End accuracy across all three environments.

\begin{minipage}{\columnwidth}

\section{Case Study}
\label{app:case-study}

\begin{tcolorbox}[
    enhanced,
    colback=white,
    colframe=black!70,
    boxrule=0.6pt,
    arc=1.2mm,
    left=1mm,
    right=1mm,
    top=5mm,
    bottom=1.2mm,
    width=\columnwidth,
    title=\textbf{FrozenLake},
    fonttitle=\bfseries\Large,
    colbacktitle=gray!12,
    coltitle=black,
    halign title=flush center,
    attach boxed title to top center={yshift=-1.6mm},
    boxed title style={
        colback=gray!12,
        colframe=black!50,
        boxrule=0.4pt,
        arc=1mm,
        left=3mm,
        right=3mm,
        top=0.6mm,
        bottom=0.6mm
    }
]

\begin{tcolorbox}[
    colback=blue!2,
    colframe=black!35,
    boxrule=0.35pt,
    arc=1mm,
    left=0.8mm,
    right=0.8mm,
    top=0.6mm,
    bottom=0.8mm,
    before skip=0pt,
    after skip=0.55em,
    title=\textbf{Input Image},
    fonttitle=\small\bfseries,
    colbacktitle=blue!8,
    coltitle=black
]
\centering
\includegraphics[width=0.75\columnwidth]{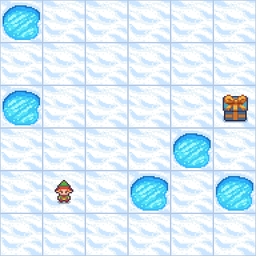}
\end{tcolorbox}

\begin{tcolorbox}[
    colback=orange!3,
    colframe=black!35,
    boxrule=0.35pt,
    arc=1mm,
    left=1.0mm,
    right=1.0mm,
    top=1.0mm,
    bottom=1.0mm,
    before skip=0pt,
    after skip=0.55em,
    title=\textbf{Question and Ground Truth},
    fonttitle=\small\bfseries,
    fontupper=\small,
    colbacktitle=orange!12,
    coltitle=black
]
\textbf{Question:} Analyze the maze and produce a safe action plan from the start to the goal without stepping into any hole.

\vspace{0.2em}
\textbf{Ground-Truth:} \textbf{URRURR}
\end{tcolorbox}

\begin{tcolorbox}[
    colback=gray!3,
    colframe=black!40,
    boxrule=0.35pt,
    arc=1mm,
    left=1.0mm,
    right=1.0mm,
    top=1.0mm,
    bottom=1.0mm,
    before skip=0pt,
    after skip=0.55em,
    title=\textbf{Qwen3-VL-4B-Instruct Reasoning Trajectory},
    fonttitle=\small\bfseries,
    fontupper=\small,
    colbacktitle=gray!12,
    coltitle=black
]
The map size is 4x4. The player starts at (3,3), and the goal is at (3,3). The H tiles are at: (0,0); (1,0); (1,3); (2,3); (3,2). The player is already at the goal, so no movement is needed. \texttt{<answer></answer>}
\end{tcolorbox}

\begin{tcolorbox}[
    colback=gray!3,
    colframe=black!40,
    boxrule=0.35pt,
    arc=1mm,
    left=1.0mm,
    right=1.0mm,
    top=1.0mm,
    bottom=1.0mm,
    before skip=0pt,
    after skip=0pt,
    title=\textbf{MGSD-4B Reasoning Trajectory},
    fonttitle=\small\bfseries,
    fontupper=\small,
    colbacktitle=gray!12,
    coltitle=black
]
The map size is 6x6. The player starts at (4,1), and the goal is at (2,5). The H tiles are at: (0,0); (2,0); (3,4); (4,3); (4,5). To safely reach the goal, the route moves upward first and then proceeds rightward through safe frozen tiles, reaching the goal at (2,5) without stepping into any hole. So the final answer is \texttt{<answer>\textbf{URRURR}</answer>}
\end{tcolorbox}

\end{tcolorbox}

\end{minipage}

\begin{tcolorbox}[
    enhanced,
    colback=white,
    colframe=black!70,
    boxrule=0.6pt,
    arc=1.2mm,
    left=1mm,
    right=1mm,
    top=5mm,
    bottom=1.2mm,
    width=\columnwidth,
    title=\textbf{Maze},
    fonttitle=\bfseries\Large,
    colbacktitle=gray!12,
    coltitle=black,
    halign title=flush center,
    attach boxed title to top center={yshift=-1.6mm},
    boxed title style={
        colback=gray!12,
        colframe=black!50,
        boxrule=0.4pt,
        arc=1mm,
        left=3mm,
        right=3mm,
        top=0.6mm,
        bottom=0.6mm
    }
]

\begin{tcolorbox}[
    colback=blue!2,
    colframe=black!35,
    boxrule=0.35pt,
    arc=1mm,
    left=0.8mm,
    right=0.8mm,
    top=0.6mm,
    bottom=0.8mm,
    before skip=0pt,
    after skip=0.55em,
    title=\textbf{Input Image},
    fonttitle=\small\bfseries,
    colbacktitle=blue!8,
    coltitle=black
]
\centering
\includegraphics[width=0.75\columnwidth]{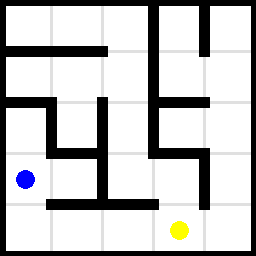}
\end{tcolorbox}

\begin{tcolorbox}[
    colback=orange!3,
    colframe=black!35,
    boxrule=0.35pt,
    arc=1mm,
    left=1.0mm,
    right=1.0mm,
    top=1.0mm,
    bottom=1.0mm,
    before skip=0pt,
    after skip=0.55em,
    title=\textbf{Question and Ground Truth},
    fonttitle=\small\bfseries,
    fontupper=\small,
    colbacktitle=orange!12,
    coltitle=black
]
\textbf{Question:} Analyze the maze image and produce a valid action plan from the player to the goal without crossing any wall.

\vspace{0.2em}
\textbf{Ground-Truth:} \textbf{LLLU}
\end{tcolorbox}

\begin{tcolorbox}[
    colback=gray!3,
    colframe=black!40,
    boxrule=0.35pt,
    arc=1mm,
    left=1.0mm,
    right=1.0mm,
    top=1.0mm,
    bottom=1.0mm,
    before skip=0pt,
    after skip=0.55em,
    title=\textbf{Qwen3-VL-4B-Instruct Reasoning Trajectory},
    fonttitle=\small\bfseries,
    fontupper=\small,
    colbacktitle=gray!12,
    coltitle=black
]

The map size is 4x4. The player starts at (3,3), and the goal is at (0,0). Open directions: (3,3): up; (0,0): right, down; (0,1): right, down; (1,0): right, down; (2,0): right, down; (3,0): up; (0,2): right, down; (1,2): right, down; (2,2): right, down; (3,2): up; (0,3): left, down; (1,3): left, down; (2,3): left, down; (3,3): up. To reach the goal, we move up from (3,3) to (2,3), then left to (2,2), then up to (1,2), then left to (1,1), then up to (0,1), then left to (0,0), and all moves follow open directions without crossing any wall. So the final answer is \texttt{<answer>\textbf{ULULU}</answer>}
\end{tcolorbox}

\begin{tcolorbox}[
    colback=gray!3,
    colframe=black!40,
    boxrule=0.35pt,
    arc=1mm,
    left=1.0mm,
    right=1.0mm,
    top=1.0mm,
    bottom=1.0mm,
    before skip=0pt,
    after skip=0pt,
    title=\textbf{MGSD-4B Reasoning Trajectory},
    fonttitle=\small\bfseries,
    fontupper=\small,
    colbacktitle=gray!12,
    coltitle=black
]
The map size is 5x5. The player starts at (4,3), and the goal is at (3,0). Open directions: (0,0): right; (0,1): right, left; (0,2): down, left; (0,3): down; (0,4): down; (1,0): right; (1,1): right, down, left; (1,2): up, down, left; (1,3): up, right; (1,4): up, down, left; (2,0): down; (2,1): up, down; (2,2): up, down; (2,3): right; (2,4): up, down, left; (3,0): up, right, down; (3,1): up, left; (3,2): up, right; (3,3): down, left; (3,4): up, down; (4,0): up, right; (4,1): right, left; (4,2): right, left; (4,3): up, right, left; (4,4): up, left. To reach the goal, we move left from (4,3) to (4,2), then left to (4,1), then left to (4,0), then up to (3,0), and all moves follow open directions without crossing any wall. So the final answer is \texttt{<answer>\textbf{LLLU}</answer>}
\end{tcolorbox}

\end{tcolorbox}

\begin{tcolorbox}[
    enhanced,
    colback=white,
    colframe=black!70,
    boxrule=0.6pt,
    arc=1.2mm,
    left=1mm,
    right=1mm,
    top=5mm,
    bottom=1.2mm,
    width=\columnwidth,
    title=\textbf{MiniBehaviour},
    fonttitle=\bfseries\Large,
    colbacktitle=gray!12,
    coltitle=black,
    halign title=flush center,
    attach boxed title to top center={yshift=-1.6mm},
    boxed title style={
        colback=gray!12,
        colframe=black!50,
        boxrule=0.4pt,
        arc=1mm,
        left=3mm,
        right=3mm,
        top=0.6mm,
        bottom=0.6mm
    }
]

\begin{tcolorbox}[
    colback=blue!2,
    colframe=black!35,
    boxrule=0.35pt,
    arc=1mm,
    left=0.8mm,
    right=0.8mm,
    top=0.6mm,
    bottom=0.8mm,
    before skip=0pt,
    after skip=0.55em,
    title=\textbf{Input Image},
    fonttitle=\small\bfseries,
    colbacktitle=blue!8,
    coltitle=black
]
\centering
\includegraphics[width=0.75\columnwidth]{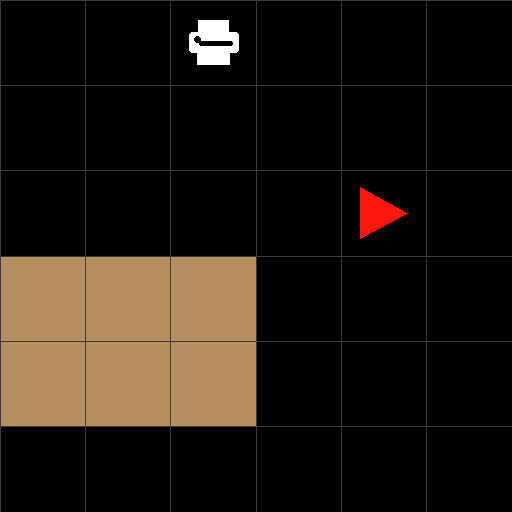}
\end{tcolorbox}

\begin{tcolorbox}[
    colback=orange!3,
    colframe=black!35,
    boxrule=0.35pt,
    arc=1mm,
    left=1.0mm,
    right=1.0mm,
    top=1.0mm,
    bottom=1.0mm,
    before skip=0pt,
    after skip=0.55em,
    title=\textbf{Question and Ground Truth},
    fonttitle=\small\bfseries,
    fontupper=\small,
    colbacktitle=orange!12,
    coltitle=black
]
\textbf{Question:} Analyze the grid image and produce a valid action plan. The agent must go to the printer, pick it up, go to the table, and drop the printer.

\vspace{0.2em}
\textbf{Ground-Truth:} \textbf{L,U,L,PICK,D,DROP}
\end{tcolorbox}

\begin{tcolorbox}[
    colback=gray!3,
    colframe=black!40,
    boxrule=0.35pt,
    arc=1mm,
    left=1.0mm,
    right=1.0mm,
    top=1.0mm,
    bottom=1.0mm,
    before skip=0pt,
    after skip=0.55em,
    title=\textbf{Qwen3-VL-4B-Instruct Reasoning Trajectory},
    fonttitle=\small\bfseries,
    fontupper=\small,
    colbacktitle=gray!12,
    coltitle=black
]
The grid size is 4x4. The agent is at (3,2), the printer is at (1,1), and the table is at (1,2). To complete the task, we move up from (3,2) to (2,2), then left to (2,1), then up to (1,1) to PICK (adjacent), then move right to (1,2) to DROP (adjacent to table). This satisfies all rules.

\texttt{<answer>\textbf{U,L,U,PICK,R,DROP}</answer>}
\end{tcolorbox}

\begin{tcolorbox}[
    colback=gray!3,
    colframe=black!40,
    boxrule=0.35pt,
    arc=1mm,
    left=1.0mm,
    right=1.0mm,
    top=1.0mm,
    bottom=1.0mm,
    before skip=0pt,
    after skip=0pt,
    title=\textbf{MGSD-4B Reasoning Trajectory},
    fonttitle=\small\bfseries,
    fontupper=\small,
    colbacktitle=gray!12,
    coltitle=black
]
The grid size is 6x6. The agent starts at (2,4), the printer is at (0,2), and the table cells are: (3,0); (3,1); (3,2); (4,0); (4,1); (4,2). To complete the task, we move up from (2,4) to (1,4), then up to (0,4) — but this is not adjacent to the printer. Instead, we must move left from (2,4) to (2,3), then up to (1,3), then left to (1,2), which is adjacent to the printer at (0,2), then PICK legally; after PICK, we move down to (2,2), which is adjacent to table cell (3,2), then DROP legally. So the final answer is \texttt{<answer>\textbf{L,U,L,PICK,D,DROP}</answer>}
\end{tcolorbox}

\end{tcolorbox}

\section{Prompts}
\label{app:prompts}

In this section, we provide the exact prompt templates used for both the privileged symbolic teacher and the visual student across our three evaluated environments: FrozenLake, Maze, and MiniBehaviour.

\begin{minipage}{\columnwidth}
\subsection{FrozenLake}

\textbf{Teacher Prompt.}~~The teacher receives the privileged symbolic context (e.g., grid size, agent position, goal location, hazard coordinates) and the reference action plan, but no image.

\begin{tcolorbox}[colframe=black!75, arc=2mm, title=FrozenLake: Teacher Prompt, fontupper=\large]
You are a FrozenLake text-only planning teacher.

You will receive a Teacher Text Context that contains a fully observable symbolic FrozenLake map, the start and goal coordinates, hole coordinates, and a reference action plan. Use the symbolic context to reason about the safe route. Do not refer to images.

Task rules:
1. The grid uses 0-based (row, column) coordinates.
2. S is the player start, G is the goal, H is a hole, and F is safe frozen land.
3. Valid actions are L (Left), D (Down), R (Right), and U (Up).
4. Moving into a hole fails. The route should finish at G.

Output requirements:
1. First, briefly state the map size, the positions of the Player and Goal, and the positions of all H tiles.
2. Next, give exactly one short sentence that states the planned route at a high level and simulate mentally and verify reaches Goal without hitting any hole.
3. Do not narrate the solution step by step. Do not list repeated moves, repeated coordinates, or intermediate states outside <answer>.
4. End with exactly one <answer>...</answer> block containing only the complete action plan, for example: <answer>LLRUD</answer>

Example Format:
The map size is 4x4. The player starts at (3,1), and the goal is at (1,2). The H tiles are at: (2,2); (2,3). To safely reach the goal, we move from (3,1) up to (2,1), then up again to (1,1), and finally right to reach the goal at (1,2), avoiding all H tiles. So the final answer is <answer>UUR</answer>
\end{tcolorbox}

\end{minipage}

\begin{minipage}{\columnwidth}

\textbf{Student Prompt.}~~In contrast to the teacher, the visual student receives only the original image and the task instruction.

\begin{tcolorbox}[colframe=black!75, arc=2mm, title=FrozenLake: Student Prompt, fontupper=\large]
You are a FrozenLake solver.

Task:
Analyze the maze and produce a safe action plan from the start to the goal without stepping into any hole.

Rules:
1. Valid actions are L (Left), D (Down), R (Right), and U (Up).
2. Frozen tiles are non-slippery.
3. Keep the reasoning brief and necessary only.
4. When mentioning positions, use 0-based (row, column) coordinates only.

Output requirements:
1. First, briefly state the map size,  the positions of the Player and Goal, and the positions of all H tiles.
2. Next, give exactly one short sentence that states the planned route at a high level and simulate mentally and verify reaches Goal without hitting any hole.
3. Do not narrate the solution step by step. Do not list repeated moves, repeated coordinates, or intermediate states outside <answer>.
4. End with exactly one <answer>...</answer> block containing only the complete action plan, for example: <answer>LLRUD</answer>

Example Format:
The map size is 4x4. The player starts at (3,1), and the goal is at (1,2). The H tiles are at: (2,2); (2,3). To safely reach the goal, we move from (3,1) up to (2,1), then up again to (1,1), and finally right to reach the goal at (1,2), avoiding all H tiles. So the final answer is <answer>UUR</answer>

Please generate the action plan for the following maze:

<TEST-IMAGE>
\end{tcolorbox}
\end{minipage}

\begin{minipage}{\columnwidth}
\subsection{Maze}

\textbf{Teacher Prompt.}~~The teacher receives the symbolic maze topology, start/end coordinates, and the reference trajectory.

\begin{tcolorbox}[colframe=black!75, arc=2mm, title=Maze: Teacher Prompt, fontupper=\large]
You are a Maze text-only planning teacher.

You will receive a Teacher Text Context that contains the maze size, start and target coordinates, an open-direction table for every cell, an optional wall-mask table, and a reference action plan. Use only this text context to solve the maze. Do not refer to images.

Task rules:
1. The grid uses 0-based (row, column) coordinates.
2. Valid actions are L (Left), D (Down), R (Right), and U (Up).
3. A move is legal only when the current cell lists that direction as open.
4. The route should finish at the target without crossing any wall.

Output requirements:
1. First, briefly state the map size, the positions of the Player and Goal, and an open-direction table for every cell.
2. Next, give exactly one short sentence that states the planned route at a high level and mentally verifies that it reaches the Goal without crossing walls.
3. Do not narrate the solution step by step. Do not list repeated moves, repeated coordinates, or intermediate states outside <answer>.
4. End with exactly one <answer>...</answer> block containing only the complete action plan using L, D, R, U, for example: <answer>DDRUR</answer>

Example Format:
The map size is 2x2. The player starts at (0,0), and the goal is at (1,1). Open directions: (0,0): right, down; (0,1): left, down; (1,0): up; (1,1): up. To reach the goal, we move right from (0,0) to (0,1), then down to (1,1), and both moves follow open directions without crossing any wall. So the final answer is <answer>RD</answer>
\end{tcolorbox}

\end{minipage}

\begin{minipage}{\columnwidth}

\textbf{Student Prompt.}~~The student receives the visual maze rendering and the standard planning prompt.

\begin{tcolorbox}[colframe=black!75, arc=2mm, title=Maze: Student Prompt, fontupper=\large]
You are a Maze solver.

Task:
Analyze the maze image and produce a valid action plan from the player to the goal without crossing any wall.

Image rules:
1. The yellow dot is the Player.
2. The blue dot is the Goal.
3. White regions are traversable corridors.
4. Black boundaries are walls and cannot be crossed.

Rules:
1. Valid actions are L (Left), D (Down), R (Right), and U (Up).
2. The player moves one grid cell per action.
3. Keep the reasoning brief and necessary only.
4. When mentioning positions, use 0-based (row, column) coordinates only.

Output requirements:
1. First, briefly state the map size, the positions of the Player and Goal, and an open-direction table for every cell.
2. Next, give exactly one short sentence that states the planned route at a high level and mentally verifies that it reaches the Goal without crossing walls.
3. Do not narrate the solution step by step. Do not list repeated moves, repeated coordinates, or intermediate states outside <answer>.
4. End with exactly one <answer>...</answer> block containing only the complete action plan using L, D, R, U, for example: <answer>DDRUR</answer>

Example Format:
The map size is 2x2. The player starts at (0,0), and the goal is at (1,1). Open directions: (0,0): right, down; (0,1): left, down; (1,0): up; (1,1): up. To reach the goal, we move right from (0,0) to (0,1), then down to (1,1), and both moves follow open directions without crossing any wall. So the final answer is <answer>RD</answer>

Please generate the action plan for the following maze:

<TEST-IMAGE>
\end{tcolorbox}

\end{minipage}

\begin{minipage}{\columnwidth}

\subsection{MiniBehaviour}

\textbf{Teacher Prompt.}~~The teacher receives the explicit state dictionary (objects, states, agent inventory) and the executable sequence of interaction actions.

\begin{tcolorbox}[colframe=black!75, arc=2mm, title=MiniBehaviour: Teacher Prompt, fontupper=\large]
You are a MiniBehaviour text-only planning teacher.

You will receive a Teacher Text Context that contains the grid size, agent position, printer position, table cells, printer-adjacent cells, table-adjacent cells, initial legal moves, a text grid, and a reference action plan. Use only this text context to solve the task. Do not refer to images.

Task rules:
1. The grid uses 0-based (row, column) coordinates.
2. Valid movement actions are L (Left), D (Down), R (Right), and U (Up).
3. The agent cannot enter any table cell. The printer cell is blocked before PICK, but after PICK the printer is removed and that cell becomes traversable.
4. PICK is legal only when the agent is adjacent to the printer.
5. DROP is legal only after PICK, when the agent is adjacent to the table.
6. The plan must perform PICK before DROP and finish immediately after a legal DROP.

Output requirements:
1. First, briefly state the grid size and the positions of the Agent, Printer, and Table.
2. Next, give exactly one short sentence that states the planned route at a high level and mentally verifies that PICK and DROP are legal.
3. Do not narrate the solution step by step. Do not list repeated moves, repeated coordinates, or intermediate states outside <answer>.
4. End with exactly one <answer>...</answer> block containing only comma-separated actions, for example: <answer>L,L,PICK,D,D,R,DROP</answer>

Example Format:
The grid size is 5x5. The agent starts at (4,0), the printer is at (3,1), and the table cells are: (1,2); (1,3); (1,4); (2,2); (2,3); (2,4). To complete the task, we move up from (4,0) to (3,0), which is adjacent to the printer at (3,1), then PICK legally; after PICK, we move right into the now-empty printer cell (3,1), then right to (3,2), which is adjacent to table cell (2,2), and DROP legally. So the final answer is <answer>U,PICK,R,R,DROP</answer>

\end{tcolorbox}

\end{minipage}

\begin{minipage}{\columnwidth}

\textbf{Student Prompt.}~~The student receives the embodied visual observation and the natural language task goal.

\begin{tcolorbox}[colframe=black!75, arc=2mm, title=MiniBehaviour: Student Prompt, fontupper=\large]
You are a MiniBehaviour solver.

Task:
Analyze the grid image and produce a valid action plan. The agent must go to the printer, pick it up, go to the table, and drop the printer.

Image rules:
1. The red object is the Agent. 
2. The white marker is the Printer.
3. The tan block is the Table.
4. Black cells are free floor cells.

Rules:
1. Valid move actions are L (Left), D (Down), R (Right), and U (Up).
2. Valid interaction actions are PICK and DROP.
3. The agent moves one grid cell per move action.
4. The agent cannot enter table cells. The printer cell is blocked before PICK, but after PICK the printer is removed and that cell becomes traversable.
5. PICK is valid only when the agent is in a cell adjacent to the printer.
6. DROP is valid only after PICK, when the agent is in a cell adjacent to the table.
7. Keep the reasoning brief and necessary only.
8. When mentioning positions, use 0-based (row, column) coordinates only.

Output requirements:
1. First, briefly state the grid size and the positions of the Agent, Printer, and Table.
2. Next, give exactly one short sentence that states the planned route at a high level and mentally verifies that PICK and DROP are legal.
3. Do not narrate the solution step by step. Do not list repeated moves, repeated coordinates, or intermediate states outside <answer>.
4. End with exactly one <answer>...</answer> block containing only comma-separated actions, for example: <answer>L,L,PICK,D,D,R,DROP</answer>

Example Format:
The grid size is 5x5. The agent starts at (4,0), the printer is at (3,1), and the table cells are: (1,2); (1,3); (1,4); (2,2); (2,3); (2,4). To complete the task, we move up from (4,0) to (3,0), which is adjacent to the printer at (3,1), then PICK legally; after PICK, we move right into the now-empty printer cell (3,1), then right to (3,2), which is adjacent to table cell (2,2), and DROP legally. So the final answer is <answer>U,PICK,R,R,DROP</answer>

Please generate the action plan for the following MiniBehaviour grid:

<TEST-IMAGE>
\end{tcolorbox}

\end{minipage}

\end{document}